\documentclass[sigconf, nonacm]{acmart}

\newcommand\vldbyear{2025}
\newcommand\vldbworkshop{1st Workshop on New Ideas for Large-Scale Neurosymbolic Learning Systems (LS-NSL)}
\newcommand\vldbauthors{\authors}
\newcommand\vldbtitle{\shorttitle} 
\newcommand\vldbavailabilityurl{https://github.com/mikl123/safe-hic}
\newcommand\vldbpagestyle{plain} 

\usepackage{xcolor} %
\usepackage{multirow} 
\usepackage{graphicx} %
\usepackage{amsmath}
\usepackage{booktabs}
\usepackage{float}
\usepackage{geometry}
\usepackage{xcolor}
\usepackage{algorithm}
\usepackage{algpseudocode}
\usepackage{paralist}
\usepackage{caption}
\usepackage{subcaption}
\usepackage{adjustbox}

\newcommand{\myeat}[1]{}

\newcommand{\myparagraph}[1]{\noindent \textbf{#1.}}

\theoremstyle{plain}
\newtheorem{definition}{Definition}

\newenvironment{example}{\refstepcounter{myexmp}\par\medskip
\noindent\textbf{Example~\themyexmp.}}{\null\hfill$\triangleleft$\medskip}
\newcounter{myexmp}[section]
\renewcommand{\themyexmp}{\thesection.\arabic{myexmp}}

\newcommand{\true}{1}
\newcommand{\false}{0}

\newcommand{\basemodel}{\textrm{B}}
\newcommand{\intmodel}{\textrm{Int}}
\newcommand{\condmodel}{\textrm{C}}
\newcommand{\sampling}{\textrm{S}}
\newcommand{\seqmodel}{\textrm{Seq}}
\newcommand{\baseseq}{\textrm{BaseSeq}}
\newcommand{\seqonly}{\textrm{SeqOnly}}

\newcommand{\base}{\textrm{Base}}
\newcommand{\ccn}{\textrm{CCN}}

\newcommand{\baseseqsat}{\textrm{BaseSeqS}}

\newcommand{\vval}{V_{\textrm{valid}}}
\newcommand{\vinval}{V_{\textrm{invalid}}}

\newcommand{\val}{\textrm{Val}}
\newcommand{\dist}{\textrm{Dist}}
\newcommand{\probass}{\textrm{PA}}
\newcommand{\ainput}{\vec i}%
\newcommand{\aval}{\vec v}%
\newcommand{\aprobass}{pa}%

\date{}

\begin{document}
\title{Constraint-aware Learning of Probabilistic Sequential Models for Multi-Label Classification}

\author{Mykhailo Buleshnyi}
\affiliation{%
  \institution{Ukrainian Catholic University}
  \city{Lviv}
  \country{Ukraine}
}
\email{buleshnyi.pn@ucu.edu.ua}

\author{Anna Polova}
\affiliation{%
  \institution{Ukrainian Catholic University}
  \city{Lviv}
  \country{Ukraine}
  }
  \email{anna.polova@ucu.edu.ua}

\author{Zsolt Zombori}
\affiliation{%
  \institution{HUN-REN Alfr\'{e}d R\'{e}nyi Institute of Mathematics}
  \institution{E{\"o}tv{\"o}s Lor\'{a}nd University}
  \city{Budapest}
  \country{Hungary}
}
\email{	zombori@renyi.hu}

\author{Michael Benedikt}
\affiliation{%
  \institution{University of Oxford}
  \city{Oxford}
  \country{UK}
}
\email{michael.benedikt@cs.ox.ac.uk}

\begin{abstract}
We investigate multi-label classification involving 
large sets of labels, where the output labels may be known to satisfy some logical constraints.  We look at an architecture in which classifiers for individual labels are fed into an expressive sequential model, which produces a joint distribution. One of the potential advantages for such an expressive model is its ability to modelling correlations, as can arise from constraints. We empirically demonstrate the ability of the architecture both to exploit constraints in training and to enforce constraints at inference time.
\end{abstract}

\maketitle
\pagestyle{\vldbpagestyle}
\begingroup\small\noindent\raggedright\textbf{VLDB Workshop Reference Format:}\\
\vldbauthors. \vldbtitle. VLDB \vldbyear\ Workshop: \vldbworkshop.\\ %
\endgroup
\begingroup
\renewcommand\thefootnote{}\footnote{\noindent
This work is licensed under the Creative Commons BY-NC-ND 4.0 International License. Visit \url{https://creativecommons.org/licenses/by-nc-nd/4.0/} to view a copy of this license. For any use beyond those covered by this license, obtain permission by emailing \href{mailto:info@vldb.org}{info@vldb.org}. Copyright is held by the owner/author(s). Publication rights licensed to the VLDB Endowment. \\
\raggedright Proceedings of the VLDB Endowment. %
ISSN 2150-8097. \\
}\addtocounter{footnote}{-1}\endgroup

\ifdefempty{\vldbavailabilityurl}{}{
\vspace{.3cm}
\begingroup\small\noindent\raggedright\textbf{VLDB Workshop Artifact Availability:}\\
The source code, data, and/or other artifacts have been made available at \url{\vldbavailabilityurl}.
\endgroup
}

\section{Introduction}

A \emph{multi-label binary classifier} is a machine learning algorithm that predicts, for each valuation of a set of input variables,  the values of multiple Boolean output variables. An \emph{output valuation} is an assignment of True or False to each of the output variables.
When the number of output valuations is small, multi-label classification can be reduced to multi-class learning: assigning to each input one of a fixed set of labels, in this case each label representing an output valuation. 
When the number of output valuations is large, we cannot perform such a reduction, due to the exponential number of combinations. 
We will consider \emph{probabilistic multi-label classification}, %
where the goal is to infer, for each valuation of a set of input variables $I_1 \ldots I_m$, a probability distribution over valuations of output variables $O_1 \ldots O_n$.
Probabilistic modeling can provide, in principle, more accurate information on the output: not just which valuation is most likely, but our degree of certainty about an individual output variable or combination of output variables.

In probabilistic modeling, there is the possibility of a high-level decision about the architecture. We call a model \emph{monolithic} if it is a single network that takes the entire input and produces a distribution on the output. We also consider a \emph{two-phase model} which starts with a \emph{single-proposition network} -- which we also call the \emph{base model} -- which has one real-valued output for each $i \leq n $, 
representing the probability for $O_i$. 
These real-valued inputs are fed into a second network, called an \emph{integrator model} that returns a distribution over output valuations.

\begin{example} \label{ex:cal500}
The CAL500 dataset \cite{call500} contains annotations of songs, with 68 input attributes and 174 output propositions that were labelled by humans. The output propositions deal with orthogonal features of a song, including whether it is a representative of a particular genre, whether or not it contains a certain kind of instrumentation, whether or not it contains female vocals, and whether it might accompany a certain kind of activity.

A specialized model could be developed for each label: for example, there might be a model meant to recognize high energy music. These models may be developed independently. On top of this we could have a model that takes as input the probabilistic classification of each individual proposition, which outputs a distribution over all propositions, or even a single most likely valuation. Intuitively, this second model would be responsible for determining the level of correlation between the different propositions on a given input. For a particular input, the specialized models might determine that it is 60\% likely that it is \textrm{angry-aggressive} and 55\% likely that it is a representative of the \textrm{folk} genre. The second model will determine that it is quite unlikely to be both.
\end{example}

Once the number of output valuations becomes large, it is difficult to design a network that outputs a distribution over it. Let us mention three options.
There is a naive approach of starting with a single-proposition network, and then calculating the probability of valuations by assuming independence~\cite{semanticloss,ccn}. Second, there is a long line of work on architectures for creating such distributions, while allowing tractable inference~\cite{tractablemodelsatlas,scalingtractable}. Finally, one can use an architecture that is extremely expressive, but where inference tasks like computing the marginal probability of a particular proposition or finding the most probable valuation are intractable.

We will focus on the last option, looking particularly at a model that reads in an output valuation as a sequence, along with the outputs of the
base model for each value in the sequence, and outputs a probability for the entire valuation. We refer to this as a \emph{sequential model}.

\begin{example} \label{ex:cal500seq} Our sequential model for the CAL500 example would take as input a sequence of 174 pairs, with each pair consisting of a valuation (true or false) for one of the output propositions, along with the probability that the base model gives to that proposition. Its output would be a probability for that valuation.
\end{example}

Giving an ordering of the output propositions as $O_1 \ldots O_n$, a \emph{prefix valuation} is an assignment of truth
values to $O_1 \ldots O_j$ for some $j < n$. A sequential model works by computing marginal probabilities for each prefix valuation, by multiplying on a new conditional probability for each $O_{j+1}$ given the marginal probability for the valuation restricted to $O_1 \ldots O_j$.

\begin{example} \label{ex:cal500seqtwo}Suppose that the first attribute is $O_1=$ \textrm{Angry-Aggressive}, one of the outputs of CAL500 and suppose the second is $O_2=$ \textrm{Emotion-Arousing-Awakening}. Consider a sequential model that is reading in a sequence starting with $O_1=\true, O_2=\false$, where the base model gives probability $0.7$ for $O_1$ and $0.6$ to $O_2$.

Based on its confidence in the base model, the sequential model, upon reading in $O_1=\true$ and $0.7$ as the base model probability, might output $0.9$ for the prefix valuation up to $O_1=\true$. Upon reading $O_2=\false$ and the base model's probability of $0.6$ for $O_2$, the sequential model will multiply the previous prefix valuation probability of $0.9$ with some number representing the conditional probability of $O_2=\false$ given $O_1=\true$. If the sequential model believes that $O_1$ and $O_2$ are dependent, it can set this conditional probability to $0.4$, leading to a prefix valuation up
to $O_1=\true, O_2=\false$ of $0.36$. 
\end{example}

A sequential model eliminates the problem of exponentially many possible outputs, since the valuation is part of the input, and there is a single output.
The drawback is that it is not obvious how to perform inference on such a model: if we want to know the probability of a particular output proposition $O_i$ being true,  computing this exactly requires aggregating over exponentially many full valuations that extend $O_i=\true$.
We address this by applying \emph{sampling-based methods} to perform approximate inference.

We mentioned that an advantage of sequential models is their ability to model correlation. In particular, they can model distributions that are consistent with known \emph{domain knowledge} in the form of \emph{constraints on the output propositions}.

\begin{example} \label{ex:cal500constraints}
In the CAL500 dataset, there are many constraints, and prior work \cite{dlwithlogicalconstraints} has extracted these.
As a trivial example, there are many cases where there are annotations for both a property and its opposite: danceable and not danceable. Clearly every song should be annotated with exactly one of these.
\end{example}

Several questions come up with respect to constraints.
Can these expressive models learn the constraints and encode them in the model?
Can these models be adapted to make use of the constraints (as ``weak supervision'') in order to learn with fewer supervised examples? 
Can these models be adapted at inference time to take into account constraints and guarantee their satisfaction?

These questions have been investigated with regard to other models in prior work, e.g. \cite{ccn}. %
Here we study them with respect to the two-phase approach, in which the second phase is a sequential model. For the first, we give empirical evidence that the models can learn constraints. For the second, we introduce an adaptation of model training to take into account constraints known to hold on unsupervised data.
And for the final item, we provide a variation of the model that can ensure enforcement of constraints.

The contributions of this paper %
are as follows:

\begin{compactitem}
   \item We propose and evaluate a two-stage architecture using a probabilistic \emph{sequential model} as the second stage for learning label distributions.

    \item We propose heuristic methods for estimating the most likely valuation.

    \item We propose additional infrastructure for enforcing constraints in our model.
    
   \item We demonstrate that using our approach we can better learn correlations in the data, including learning relationships corresponding to constraints: thus we do not need to make use of domain knowledge, we can often learn it.

\end{compactitem}

\section{Related Work}

\myparagraph{Multi-Label Classification}
The problem of classifying data with multiple, potentially overlapping or correlated, labels has received considerable study. One obvious approach is to assume independence of the different labels, which limits expressiveness but reduces to a single-classification problem. Another approach models limited correlation -- for example, pairwise correlations between labels.
See, for example, the discussion in \cite{multilabelprior}.

\myparagraph{Probabilistic circuits}
A long line of research stemming from the theory community studies probabilistic models over a large output space, where inference is still tractable: see, e.g. \cite{tractablemodelsatlas, scalingtractable}. Models are given in the form of circuits where leaves represent some tractable atomic model and gates represent simple operations such as products. Restrictions on the structure of the circuit are imposed to achieve tractability.

\myparagraph{Transformers for completing data}
Expressive models such as transformers have been applied to almost every possible data-related task. Probably closest to our work is inferring
missing data in tables, which for categorical data can be seen as multi-label prediction. See, e.g. \cite{imputation} for an overview.
The common assumption is that most of the values within a table, even in a particular row, are present, and the goal is to infer the missing ones;  In our work we focus on predicting a full set of labels. While work on imputation is usually end-to-end, we focus on a two-phase architecture working on top of predictors for individual labels.

\myparagraph{Integrating Probabilistic Classifiers}
We consider a two-phase model where the second model reconciles probabilities returned from an initial model or models for individual propositions.  The initial model could thus be considered as an estimate of marginal probabilities for the joint distribution returned by the second model. The problem of computing a joint distribution for a set of outputs, consistent with a set of marginals for each output, has been studied in the past. It is known to be hard: see, for example \cite{marginals}.
For integrating the predictions of the initial model, we use a sequential approach, which is known as \emph{next token prediction} in the language modeling literature \cite{nexttokenprediction}.

\myparagraph{Learning with Constraints}
Learning in the presence of domain constraints has been studied from many perspectives. From the point of view of \emph{training}, one question is how to make use of the constraints on unsupervised data, as a kind of weak supervision.
In addition, or in lieu of this, one can make use of constraints at \emph{inference time}, adjusting the inference procedure so that it always produces constraint-satisfying outputs.
\cite{learningvsrepairingwithconstraints} studies both aspects of the problem from the point of view of sample complexity and accuracy of learning, dealing with a simple model of constraints given as a disjunction of values for each input. For constraints given as propositional formulas on the output space, \cite{dlwithlogicalconstraints,ccn} propose algorithms for both training and inference. In our work, we consider constraints of this form, comparing our results with those in \cite{ccn}: see Section~\ref{sec:experiments} for more detail on \cite{ccn}.

We deal with input-independent propositional logic constraints on the output labels. A very different kind of constraint comes from scenarios where a learner observes information derived from a set of hidden labels. For example, a learner might be interested in classifying a sequence of images of digits, and observe only their sum \cite{deepproblog}. Thus each observation represents a disjunction of possibilities for the desired function outputs, thus as a kind of sample-dependent constraint. Learning algorithms for this scenario have received significant attention both in theory and practice, most recently in \cite{neuroconstraints1, neuroconstraints2}. A related ``dynamic constraint'' scenario, motivated by learning in the presence of rules, is discussed
in \cite{towardsunbiased}.

\section{Two-Stage Inference Using a Sequential Model}
\label{sec:sequential}

We introduce a two-stage learning architecture. In the first stage, a feedforward neural network, which we will refer to as the \emph{base model} processes the input and outputs a marginal probability prediction for each output label. In the second stage, an \emph{integrator model} is used, which combines marginal probabilities into a prediction of the probability of an entire output valuation.

We now describe the architecture more formally. We fix input variables $\vec I=I_1 \ldots I_m$, along with output propositions $O_1 \ldots O_n$. We let
$\val(\vec I)$ ($\val(\vec O)$) be the set of assignments to $\vec I$ ($\vec O$). To simplify the terminology, we refer to an input assignment $\vec i \in \val(\vec I)$ as an \emph{input} and an output assignment $\aval \in \val(\vec O)$ as a \emph{valuation}.
We let $\dist(\val(\vec O))$ be the set of \emph{output joint distributions}: probability distributions over $\val(\vec O)$.
An \emph{output marginal probability assignment} \ is an assignment
of each $O_i$ to a probability. We let $\probass(\vec{O})$ be
the set of such assignments.

\begin{definition}[Base Model]
\label{def:base}
A \emph{base model} $\basemodel_\theta$ is a class of functions, parameterized by trained parameters $\theta$, that maps each input assignment to an output marginal probability assignment. Its signature is $\val(\vec I) \rightarrow \probass(\vec O)$.
\end{definition}

The base model's prediction can be interpreted as a probability distribution over valuations by assuming that the output variables are independent. With this assumption, the probability of valuation $\aval$ for input $\ainput$ is 
$$P_{\basemodel}(\ainput, \aval) = \prod_{1 \leq j \leq n} \left( v_j \basemodel(\ainput)_j + (1-v_j) (1-\basemodel(\ainput)_j) \right)$$.
\begin{definition}[Integrator Model]
An \emph{integrator model} is any function $\intmodel_{\theta'}$ parametrized by $\theta'$ that takes as input a marginal probability assignment and returns an output joint distribution. Its signature is $\probass(\vec O) \rightarrow \dist(\val(\vec O))$.
\end{definition}

A \emph{prefix valuation} for $j \leq n$ is a valuation on $O_1 \ldots O_j$. For a full output valuation $\aval \in \{0,1\}^n$ and $j \leq n$, we let $\aval|j \in \{\false,\true\}^j$ be the prefix valuation obtained by restricting $\aval$ to $O_1 \ldots O_j$. We focus on integrator models that rely on the repeated invocation of \emph{prefix conditional models}:

\begin{definition}[Prefix Conditional Model]
A \emph{Prefix conditional model} is any function $\condmodel_{\theta'}$ parameterized by $\theta'$ that takes as input a marginal probability assignment and a prefix valuation and returns a value in $[0,1]$ interpreted as the probability of the output variable immediately following the prefix being true. Its signature is $(\probass(\vec O), \val(\vec O)) \rightarrow \mathbb{R}$. 
\end{definition}

A prefix conditional model $\condmodel$ implicitly implements an integrator model 
as it defines a distribution over $\val(\vec O)$, given marginal probability assignment $\aprobass \in \probass(\vec O)$: the probability of any valuation $\aval \in \val(\vec O)$ is given by 
\begin{align*}
P_{\condmodel}(\aprobass, \aval)
& = \prod_{0 \leq j < n} \left[ v_{j+1} \condmodel(\aprobass, \aval|j) + (1-v_{j+1}) (1-\condmodel(\aprobass, \aval|j))\right] 
\end{align*}
\begin{definition}[Sequential Model]
An integrator model 
that relies on the repeated invocation of prefix conditional model $\condmodel$ is called a \emph{Sequential model} $\seqmodel_{\condmodel}$.
\end{definition}

In this paper we are interested in training a two-phase architecture where the integrator is a sequential model. 

\begin{definition}[Base-Seq Model]
\label{def:baseseq}
    A two phase architecture $\baseseq = (\basemodel, \seqmodel_{\condmodel})$ is a 
    predictor for multi label classification that uses base model $\basemodel$ to map inputs to marginal probability predictions and sequential model $\seqmodel_\condmodel$ to generate a distribution on $\val(\vec O)$.
\end{definition}

One could also consider using a sequential model in itself, i.e., without a base model. In this case, the sequential model takes the initial input, instead of the marginal probability predictions. We discuss this option in the appendix only.

\subsection{Inference with a Sequential Model}

Evaluating a sequential model $\seqmodel_{\condmodel}$ on a single valuation requires $n$ invocations of $\condmodel$. To compute the full distribution on $\val(\vec O)$ we have to repeat this for all valuations, which requires $n 2^n$ model evaluations. This is prohibitively expensive for larger $n$. 

In practice, we often do not need the full distribution. Instead, we are content with identifying the most probable valuations. Unfortunately, this too, is known to be intractable, 
however, effective approximation algorithms exist based on some \emph{sampling strategy}.

\begin{definition}[Sampling Strategy]
    By \emph{sampling strategy} we refer to any (possibly non deterministic) function $\sampling$ that receives a probability estimate and outputs a binary valuation. %
\end{definition}

A sequential model $\seqmodel_{\condmodel}$ and a sampling strategy $\sampling$ can be jointly used to obtain a sample from $\dist(\val(\vec O))$ by starting from the empty prefix valuation and iteratively extending it based on $\sampling$ applied to the output of $\condmodel$ on the prefix, as described by Algorithm~\ref{alg:seqsampling}.
    \begin{algorithm}
        \hspace*{\algorithmicindent} \textbf{Input} $\aprobass \in \probass(\vec O)$
        \begin{algorithmic}
            \State $\aval \gets ()$
            \While {$|\aval| < n$}
            \State $i \gets |\aval|+1$
            \State $P_i \gets \condmodel(\aprobass, \aval)$
            \State $v_i \gets \sampling(P_i)$
            \State $\aval \gets append(\aval, v_i)$
            \EndWhile
        \end{algorithmic}
        \hspace*{\algorithmicindent} \textbf{Output} $\aval \in \val(\vec O)$
        \caption{Selecting a sample from $\dist(\val(\vec O))$ based on sequential model $\seqmodel_{\condmodel}$ and sampling strategy $\sampling$.}
        \label{alg:seqsampling}
    \end{algorithm}  
We can think of the inference process as a decision tree. In each step the model makes a decision whether the current output variable is true or false. These decisions are accumulated in the prefix valuation that forms the input of the prefix conditional model.

One popular generalization of Algorithm~\ref{alg:seqsampling} is \emph{beam search}~\cite{beamsearch,beamsearch_ml}, 
which tries to heuristically identify some of the most probable valuations. Starting from the empty prefix, it iteratively extends the considered prefixes. However, rather than exploring all prefix valuations, as in a brute-force exploration, beam search keeps only a fixed $k$ number of most likely prefixes (according to the prefix conditional model) where $k$ is called the \emph{beam width}, thus reducing computational complexity. At each iteration, $k$ stored prefixes are extended into $2k$ possible continuations of which the $k$ most probable are selected for the next iteration. Eventually, beam search outputs $k$ full valuations, the most probable of which can be considered as the model's output.
Using this approach, we cannot guarantee either that the model will predict a valid combination or that it will predict the most likely combination, but if we use a large beam width we can expect the model to find something valid and of high probability.

\subsection{Supervised Learning with Base-Seq Model}

Given supervised labels, a base-seq model $\baseseq = (\basemodel, \seqmodel_{\condmodel})$ can be trained to minimize the negative log likelihood of the target valuation. Given input $\ainput$ and target valuation $\aval$, the loss is 
\begin{align*}
    & \mathcal{L}_{\textrm{sup}}(\aval) = - \log(P_{\condmodel}(\aprobass, \aval)) \\
    & \mbox{ where } \aprobass = \basemodel(\ainput) \\
    & \mbox{ and } \log(P_{\condmodel}(\aprobass, \aval)) = \\
    & \log \left[\prod_{0 \leq j < n} \left( v_{j+1}  \condmodel(\aprobass, \aval|j) + (1-v_{j+1}) (1-\condmodel(\aprobass, \aval|j) \right) \right] \\
    & = \sum_{0 \leq j < n} \log\left( v_{j+1}  \condmodel(\aprobass, \aval|j) + (1-v_{j+1}) (1-\condmodel(\aprobass, \aval|j) \right)     
\end{align*}

\subsection{Unsupervised Learning with Base-Seq Model} \label{subsec:unsupervised}

We present two possible ways learning from unsupervised data in the presence of constraints.

\myparagraph{Pseudo Labeling}
\emph{Pseudo labeling} was introduced in \cite{lee2013pseudo}. It assumes that we have both a supervised and an unsupervised dataset. The model is first trained on the supervised data. After convergence, the model -- combined with beam search -- is used to evaluate each unsupervised input $\ainput$ once 
and get an estimate of the most likely output valuation that is constraint consistent. This valuation is assigned to $\ainput$ as the pseudo label. If beam search returns no valid valuation, then the given input is discarded. Once pseudo labels are assigned, another round of training is performed, this time using both the supervised and the unsupervised samples. 

\myparagraph{Constraint loss}
Our second approach is based on the additional training signal provided by constraints.
Prior works, e.g. \cite{semanticloss,deepproblog} compute the total probability of constraint violation, which however, is not tractable with our model. Hence we resort to sampling.
Given an input $\ainput$ and a base-seq model $\baseseq = (\basemodel, \seqmodel_{\condmodel})$, we can use beam search to obtain $k$ output valuations $V = \aval^1 \ldots \aval^k$ that the model considers to have high probability. We can split them into constraint consistent valuations $\vval$ and constraint violating ones $\vinval$. Naturally, the probability of the invalid valuations should be minimized. We can minimize the log probability of the valuations in $\vinval$:
\begin{align*}
    & \mathcal{L}_{\textrm{cons}}(V) = \sum_{\aval \in \vinval} \log(P_{\condmodel}(\aprobass, \aval))
\end{align*}
The loss function above will penalize every decision that leads to the final valuation. This, however, can be contra-productive: If the true valuation shares a prefix with an inconsistent valuation, then the true valuation can become less probable. This could be avoided by only penalizing decisions that are not on some prefix of the true valuation. However, since we do not assume access to the true valuation during training, we propose to approximate this with the constraint consistent samples $\vval$ obtained from beam search: whenever some prefix of a sampled invalid valuation is also part of a sampled valid valuation, then the probability of the prefix is not altered, only that of the subsequent steps.

For any $\aval \in \vinval$ and prefix $\aval|j$ let $m_{\aval|j} \in \{0,1\}$ be a  variable that is zero if there exists some valid sampled valuation $\aval' \in \vval$ and prefix $\aval'|j'$ such that $\aval|j = \aval'|j'$ and $1$ otherwise. 
The value of the revised log probability, $\log( P'_{\condmodel}(\aprobass, \aval) )$, is now %
\begin{align*}
    \sum_{0\leq j < n}   m_{\aval|j+1} \log [  v_{j+1}  \condmodel(\aprobass, \aval|j)  + (1-v_{j+1}) (1-\condmodel(\aprobass, \aval|j) ] 
\end{align*}

\begin{example} 
\label{ex:sampling} 
Suppose that we have three output variables $n=3$. Let us assume that sampling returns two valid and two invalid valuations:
\begin{align*}
    \vval & = \{(\false, \false, \false), (\true, \true, \false)\} &
    \vinval & = \{(\false, \false, \true), (\true, \false, \true) \} \\
    V & = \vval \bigcup \vinval
\end{align*}
The first invalid valuation shares a two step prefix with the first valid one, hence only its third prediction should be penalized. Similarly, the second invalid valuation shares a one step prefix with the second valid validation, hence only its last two predictions are penalized. The loss  $\mathcal{L}_{\textrm{cons}}(V)$  is the sum of the log probabilities of these three predictions:
\begin{align*}
    \log(\condmodel(\aprobass, (\false, \false))) %
    + (1 - \log(\condmodel(\aprobass, (\true)))) %
    + \log(\condmodel(\aprobass, (\true, \false)))
\end{align*}
\end{example}

The constraint loss $\mathcal{L}_{\textrm{cons}}$ can be added to the supervised loss $\mathcal{L}_{\textrm{sup}}$ to put more emphasis on reducing the predicted probability of invalid valuations. Furthermore, it can also be used as a single loss term when only unsupervised data is available. In many real-world applications, labeled data is scarce and expensive, while it is much easier to access large amounts of unsupervised samples.

\subsection{Enforcing Safety in the Base-Seq Model}

Optimizing constraint loss $\mathcal{L}_{\textrm{cons}}$ can reduce the likelihood of the model giving an invalid answer, but it gives no guarantees of full satisfaction. To provide such a guarantee, we incorporate an additional safety mechanism into the inference via integrating a SAT solver. At each step of the beam search, the SAT solver checks if the prefix valuation associated with each beam is satisfiable with respect to the provided constraints. Unsatisfiable prefixes are eliminated.

The satisfiability of propositional formulas is an NP-hard problem, however, modern SAT solvers are well developed to handle large problems. Furthermore, this approach enables us to enforce any logical constraints unlike \cite{ccn}.

\section{Experiments}
\label{sec:experiments}

We aim to investigate how learning can benefit from a sequential approach in constrained scenarios. In exploring this, we conduct a series of experiments
aimed at addressing the following questions:

\begin{compactitem}
    \item What is an efficient method for identifying the most probable combination in a joint distribution? How does beam width influence model performance?
    \item How can constraints be effectively incorporated into sequential model training?
    \item How does the sequential model perform in unsupervised
    settings?
\end{compactitem}

\paragraph{Metrics}
Our inference procedures return either a single valuation (the default below) or a set of such valuations. We use the following metrics in our research to evaluate the performance of our models:

\noindent \textbf{Accuracy}: The fraction of our predicted valuations that exactly match the ground truth.

\noindent \textbf{Top-k Accuracy}: The fraction of completely correctly predicted valuations within the top \( k \) predicted valuations. This is applicable only when we use a beam search that returns multiple answers.

\noindent \textbf{Constraint violation}: The fraction of predicted valuations that violate at least one constraint.

\noindent \textbf{Target probability}: Joint probability of target valuation.

\paragraph{Models}
We present a series of experiments comparing learning architectures introduced in Section~\ref{sec:sequential}. Unless otherwise specified, we use the following model architectures:

\noindent $\base$:
As base model, (see Definition~\ref{def:base}), we use the same architecture as in \cite{ccn}: a neural network with two hidden layars and a Sigmoid activation at the final layer to normalize values into $[0,1]$. The number of neurons in the hidden layer is dataset specific and is given in Table~12 of \cite{ccn}. The final layer is preceded by a dropout layer during training.

\noindent $\baseseq$:
Our common Base model is extended with a sequential integrator (see Definition~\ref{def:baseseq}), which is a two layer neural network with 300 hidden units. 
    
\noindent $\baseseqsat$:
This is the same as $\baseseq$ but it guarantees constraint satisfaction by using a SAT solver on each prediction step. In our implementation, we use PySAT~\cite{pysat}.

\noindent $\ccn$:
Our common Base model is extended with a non-parametric 
layer, as described in \cite{ccn}.  This layer is specific to a set of constraints, roughly speaking propositional Horn clauses. It adjusts the truth values (taking min or max of certain labels) to ensure a constraint-consistent final prediction. The key strength of this approach is that the $\ccn$-layer is used not only for inference, but also during training: the gradient of the final loss is backpropagated through the $\ccn$-layer. On the other hand, it only works for restricted cases, and makes heuristic choices with respect to how constraint violating predictions are fixed. Both of these restrictions are lifted in the a Base-Seq model. We use $\ccn$ as our primary baseline, since the restrictions are satisfied for all of the constraints we use.

\paragraph{Training Hyperparameters}
For $\base$ and $\ccn$ we use the same configuration as in \cite{ccn}. We use the Adam optimizer, learning rate $0.0001$, weight decay $0.0001$,  dropout rate $0.8$, batch size $4$. We employ early stopping with a patience of $20$ epochs. The model uses 2 hidden layers and the number of hidden dimensions depends on the dataset, see Table~12 of \cite{ccn}. 

For the sequential model in $\baseseq$ and $\baseseqsat$ we provide the default training configurations and only indicate in the experiment descriptions when there is a deviation from these. We use the Adam optimizer, learning rate $0.001$. weight decay $0.001$,  dropout rate $0.1$, batch size $16$. We employ early stopping with a patience of $20$ epochs. Beam search based inference uses beam width $4$. The model has 2 hidden layers and uses hidden dimension $300$.

\subsection{Comparison of the Sequential Architecture with Baselines}
For the reader who wants some additional intuition for the comparison between our sequential  model and $\ccn$, we step through a toy example in Appendix \ref{app:toyexample}. In the body of the paper we begin
with more realistic problems, specifically the  datasets introduced in \cite{ccn}. These datasets vary significantly: the number of output variables ranges from 6 to 174, and the number of constraints spans from 1 to 344. For consistency, we replicate the experiments reported in \cite{ccn} and present our own results. 

\begin{table}[htb]
\caption{Comparing accuracy values of trained Base, Base-seq, Base-seqS and $\ccn$ models on 11 datasets from \cite{ccn}.}
\label{tab:approach_comparison}
\centering
\addtolength{\tabcolsep}{-0.5em}
\begin{tabular}{lcccccc}
\toprule
\textbf{Model} & \textbf{Emotions} & \textbf{Yeast} & \textbf{Arts} & \textbf{Cal500} & \textbf{Enron} & \textbf{Genbase} \\
\midrule
Base           & 0.264 & 0.185 & 0.209 & \textbf{0.000} & 0.122 & 0.984 \\
Base-seq       & \textbf{0.349} & \textbf{0.235} & \textbf{0.365} & \textbf{0.000} & \textbf{0.151} & 0.985 \\
Base-seqS      & \textbf{0.349} & \textbf{0.235} & \textbf{0.365} & \textbf{0.000} & \textbf{0.151} & 0.985 \\
$\ccn$         & 0.311 & 0.176 & 0.222 & \textbf{0.000} & 0.140 & \textbf{0.988} \\\\
\midrule
\midrule
\textbf{Model} & \textbf{Image} & \textbf{Medical} & \textbf{Scene} & \textbf{Science} & \textbf{Business} \\
\midrule
Base           & 0.428 & 0.515 & 0.618 & 0.204 & 0.580 \\
Base-seq       & 0.501 & 0.515 & \textbf{0.700} & \textbf{0.341} & \textbf{0.589} \\
Base-seqS      & \textbf{0.502} & 0.515 & \textbf{0.700} & \textbf{0.341} & \textbf{0.589} \\
$\ccn$         & 0.490 & \textbf{0.538} & 0.652 & 0.208 & 0.583 \\
\bottomrule
\end{tabular}
\end{table}

\begin{table}[htb]
\caption{Comparing target probability values of trained Base, Base-seq, and $\ccn$ models on 11 datasets.} 
\label{tab:target_prob_comparison}
\centering
\addtolength{\tabcolsep}{-0.5em}
\begin{tabular}{lccccccc}
\toprule
\textbf{Model} & \textbf{Emotions} & \textbf{Yeast} & \textbf{Arts} & \textbf{Cal500} & \textbf{Enron} & \textbf{Genbase} \\
\midrule
Base & 0.138 & 0.044 & 0.106 & \textbf{0.000} & 0.043 & 0.950 \\
Base-seq & \textbf{0.145} & \textbf{0.131} & \textbf{0.211} & \textbf{0.000} & \textbf{0.055} & \textbf{0.976} \\
$\ccn$     & 0.142 & 0.043 & 0.107 & \textbf{0.000} & 0.048 & 0.951 \\
\midrule
\midrule
& \textbf{Scene} & \textbf{Science} & \textbf{Business} & \textbf{Image} & \textbf{Medical} \\
\midrule
Base & 0.492 & 0.140 & 0.397 & 0.282 & 0.402 \\
Base-seq & \textbf{0.608} & \textbf{0.212} & \textbf{0.467} & \textbf{0.386} & 0.355 \\
$\ccn$     & 0.476 & 0.139 & 0.393 & 0.279 & \textbf{0.423} \\\\
\bottomrule
\end{tabular}
\end{table}

We train the $\base$, $\baseseq$, $\baseseqsat$ and $\ccn$ models on these 11 datasets and report final accuracies in Table~\ref{tab:approach_comparison}. 
We see little variance on two datasets: 1) Cal500 is too hard for all models and they achieve 0\% accuracy and 2) Genbase is quite easy and all models are close to being perfect. On the remaining nine datasets, the $\baseseq$ model performs remarkably well, despite the fact that it makes no use of the constraints either during training or during inference. The $\baseseq$ model outperforms both the Base model and $\ccn$ on 8 of the 9 datasets, often by a large margin. For example, $\baseseq$ reaches $36\%$ on Arts, compared to 21\% and 22\% for $\base$ and $\ccn$, respectively. Note furthermore, adding a SAT solver barely ever changes the model's accuracy, indicating that the $\baseseq$ model in itself is capable of extracting and learning constraints directly from data. 

The only dataset where $\ccn$ performs significantly better is Medical ($53.8\%$ vs $51.5\%$), which has very simple constraints of the form $O_i \implies O_j$. For such constraints, we hypothesize that the ordering of the variables in the base-seq model can play an important role: it is better to first predict subclasses and then  superclasses. Indeed, preliminary experiments suggest that changing the variable ordering on this dataset can result in $1-2\%$ accuracy changes. However, considering all datasets,  the effect of variable ordering is less clear and we leave this topic for future work.

Table \ref{tab:target_prob_comparison} shows the probability of the true valuation for the different competitors. We see that Base-Seq gives the highest probability on most entries.

\subsection{Top-k Accuracy on the Full Distribution}
The previous evaluations used metrics based on a single predicted valuation. This can be misleading when the output space is large: in such scenarios, even if the most likely prediction is not the true one, we might still consider the model to perform well if the target valuation is among the top k predictions. In this experiment we explore top-k accuracy. We identify the top-k predictions based on the full output distribution, which however, can only be computed when the number of output variables is small. We perform this experiment on \textbf{Yeast}, \textbf{Scene} and \textbf{Emotions} datasets that have 14, 6 and 6 output variables, respectively.
Table~\ref{tab:topk_accuracy} shows top-k accuracies. For all k values, the $\baseseq$ model outperforms both the $\base$ model and $\ccn$ by a large margin.

\begin{table}[htb]
\caption{Top-$k$ accuracy comparison between CCN, Base, and Base-Seq models across different datasets.}
\label{tab:topk_accuracy}
\centering
\addtolength{\tabcolsep}{-0.1em}
\begin{tabular}{l|l|c c c c}
\toprule
Dataset & Model & Top-1 & Top-2 & Top-5 & Top-10 \\ 
\midrule
\multirow{4}{*}{Emotions} 
    & CCN                & 0.311 & 0.441 & 0.624 & 0.786 \\
    & Base               & 0.264 & 0.396 & 0.655 & 0.821 \\ 
    & Base-seq (beam)    & \textbf{0.349} & \textbf{0.514} & \textbf{0.750} & \textbf{0.863} \\ 
\midrule
\multirow{4}{*}{Yeast} 
    & CCN                & 0.176 & 0.230 & 0.293 & 0.346 \\
    & Base               & 0.185 & 0.228 & 0.289 & 0.341 \\ 
    & Base-seq (beam)    & \textbf{0.235} & \textbf{0.314} & \textbf{0.452} & \textbf{0.568} \\ 
\midrule
\multirow{4}{*}{Scene} 
    & CCN                & 0.652 & 0.733 & 0.870 & 0.972 \\
    & Base               & 0.618 & 0.742 & 0.937 & 0.991 \\ 
    & Base-seq (beam)    & \textbf{0.700} & \textbf{0.840} & \textbf{0.946} & \textbf{0.993} \\ 
\bottomrule
\end{tabular}
\end{table}

\subsection{The Impact of Beam Search}
Next, we contrast top-k predictions based on the full distribution with its approximation based on beam search, using the \textbf{Yeast}, \textbf{Scene} and \textbf{Emotions} datasets. The results are summarized in Table~\ref{tab:beamsearch}. We observe almost identical top-k accuracy scores, with the only differences in top-10 accuracy. This suggests that beam search returns a close approximation of the most likely valuations.

\begin{table}[htb]
\caption{Accuracy comparison of the Base-Seq model based on computing the full distribution (Exact) and on beam search with beam width 10.}
\label{tab:beamsearch}
\centering
\addtolength{\tabcolsep}{-0.1em}
\begin{tabular}{l|l|c c c c}
\toprule
Dataset & Model & Top-1 & Top-2 & Top-5 & Top-10 \\ 
\midrule
\multirow{2}{*}{Emotions} 
    & Base-seq (beam) & 0.349 & 0.514 & 0.750 & 0.863 \\ 
    & Base-seq (exact)  & 0.349 & 0.514 & 0.750 & 0.864 \\ 
\midrule
\multirow{2}{*}{Yeast} 
    & Base-seq (beam) & 0.235 & 0.314 & 0.452 & 0.568 \\ 
    & Base-seq (exact)  & 0.235 & 0.314 & 0.452 & 0.567 \\ 
\midrule
\multirow{2}{*}{Scene} 
    & Base-seq (beam) & 0.700 & 0.840 & 0.946 & 0.993 \\ 
    & Base-seq (exact)  & 0.700 & 0.840 & 0.946 & 0.993 \\ 
\bottomrule
\end{tabular}
\end{table}

\subsection{Impact of Beam Width}
As previously mentioned, finding the most probable valuation is intractable in a base-seq model. Thus we employ an approximation based on beam search. %
In beam search, the choice of the beam width $k$ provides an interpolation between fully greedy search ($k=1$) and computing the entire distribution ($k=\infty$).

\begin{figure}[htb]
    \centering
    \includegraphics[width=0.98\linewidth]{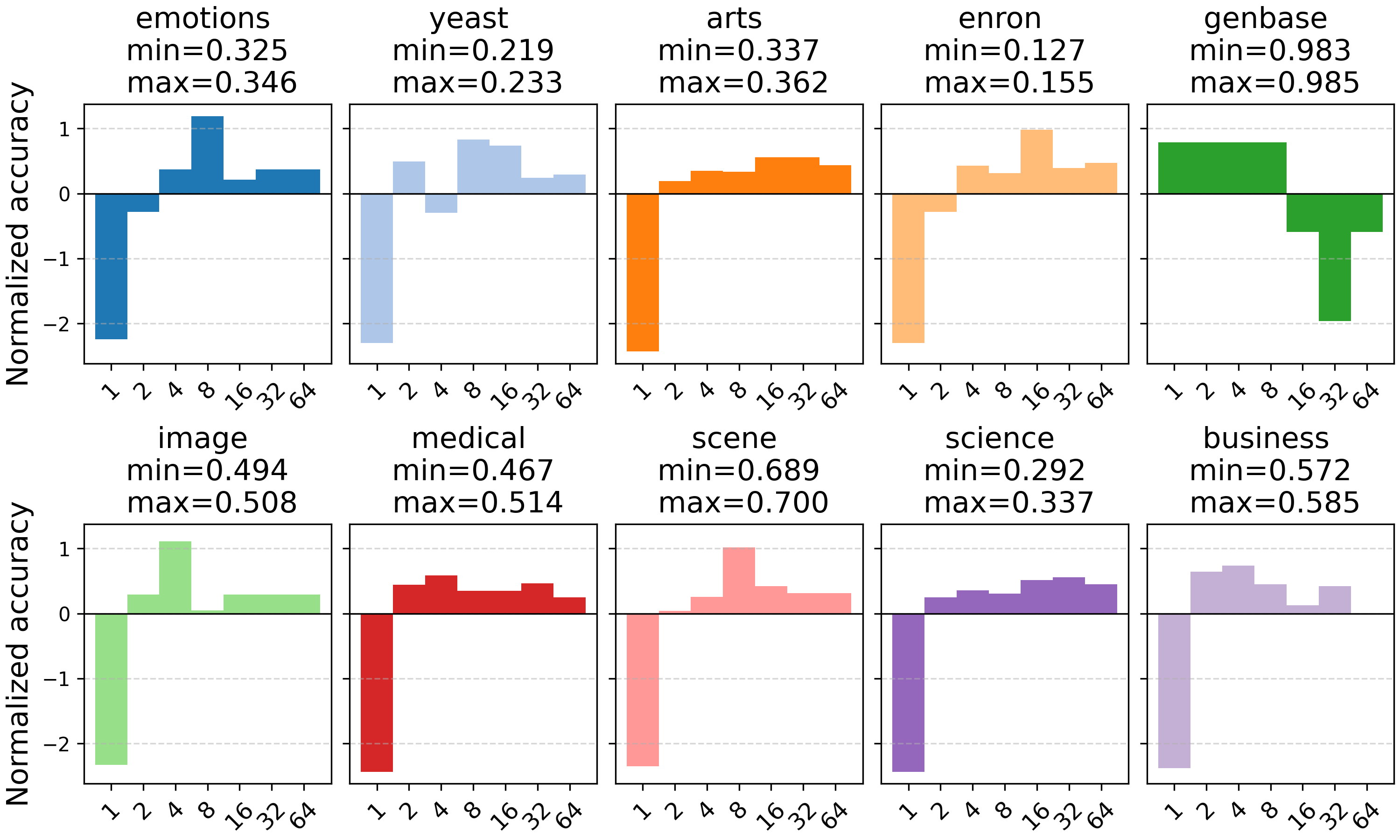}
    \caption{Normalized accuracy for each beam width and dataset. The normalization is performed according to the formula:
    $ \text{Normalized Accuracy} = \frac{acc - \mu}{\sigma} $, where $ \mu $ and $ \sigma $ denote the mean and standard deviation of the averaged accuracies, respectively. Minimum and maximum accuracies are specified in the labels of each image. Each experiment was repeated 5 times and average accuracies are reported.}
    \label{fig:beamsearch}
\end{figure}

In this experiment we evaluate the model's sensitivity to beam width. For each dataset, we evaluate the trained model with beam width $k \in \{1, 2, 4, 8, 16, 32, 64\}$. The results are shown in Figure~\ref{fig:beamsearch}. We find that fully eager search ($k=1$) is significantly weaker than for larger $k$ values, so there is a clear benefit of using a non trivial beam width. However, performance plateaus near beam width $4$.

\subsection{Constraint Satisfaction}
$\ccn$ and $\baseseqsat$ explicitly enforce all constraints, while Base and Base-Seq provide no guarantees. We now examine how often the latter two models violate constraints.
Table~\ref{tab:constraint_satisfaction} shows constraint violation ratio on the 11 datasets from \cite{ccn}. We note that in many cases the base model is already capable of fully satisfying constraints. Exceptions are the Emotions, Yeast, Scene and Image datasets with over 10\% violation. For  these datasets, the base-seq model reaches zero or close to zero constraint violation, demonstrating the architecture's ability to internalize constraints directly from the data.

\begin{table}[htb]
\caption{Constraint violation ratio of trained Base and Base-Seq models on 11 datasets from \cite{ccn}.}
\label{tab:constraint_satisfaction}
\centering
\addtolength{\tabcolsep}{-0.5em}
\begin{tabular}{lccccccc}
\toprule
\textbf{Model} & \textbf{Emotions} & \textbf{Yeast} & \textbf{Arts} & \textbf{Cal500} & \textbf{Enron} & \textbf{Genbase} \\
\midrule
Base & 0.107 & 0.101 & \textbf{0.000} & \textbf{0.000} & \textbf{0.000} & 0.001 \\
Base-Seq & \textbf{0.000} & \textbf{0.000} & \textbf{0.000} & 0.036 & \textbf{0.000} & \textbf{0.000} \\
\midrule
\midrule
& \textbf{Scene} & \textbf{Science} & \textbf{Business} & \textbf{Image} & \textbf{Medical} \\
\midrule
Base & 0.167 & \textbf{0.000} & \textbf{0.000} & 0.200 & 0.001 \\
Base-Seq & \textbf{0.000} & 0.001 & \textbf{0.000} & 0.001 & \textbf{0.000} \\
\bottomrule
\end{tabular}
\end{table}

\myeat{
\subsection{Impact of Order}

The sequential nature of the integrator model implies that the order in which output propositions are processed can have a substantial impact on the final output distribution. Intuitively, presenting output propositions that are easier to predict or more informative earlier in the sequence can guide the conditional model to make more accurate predictions for subsequent propositions, thereby improving overall performance.

To investigate, we explored various orderings of the output propositions:
\begin{inparaenum}
\item \textbf{Reversed Order:} Inverting the default ordering to examine whether processing direction influences performance.
\item \textbf{Class Balance Order:} Ordering by the balance of label distributions in the dataset, based on the hypothesis that more balanced classes may provide stronger and more stable predictive signals for later decisions.
\item \textbf{Base Model Accuracy Order:} Sorting propositions according to the accuracy of the base model’s marginal probability predictions, so that more reliable propositions are processed first by the integrator model.
\item \textbf{Hierarchy Order:} For datasets with a known class hierarchy in the constraints, ordering according to their level of generality to respect natural semantic groupings.
\end{inparaenum}
The results (see App. ~\ref{app:ordering} for details) indicate that no single ordering outperforms the others across datasets, despite the fact that ordering can have a large impact for each dataset.
}%

\subsection{Exploiting Unsupervised data}
In this experiment, we evaluate the two approaches outlined in Section \ref{subsec:unsupervised} for leveraging unsupervised data: pseudo labeling and constraint loss. Starting from the original train/validation/test split, we further divide the training set into supervised and unsupervised portions, based on a specified ratio. For example, a value of \(0.3\) indicates that \(30\%\) of the training data is treated as unsupervised, while the remaining \(70\%\) is used as supervised data.

We use a beam width of $5$ for training. For pseudo labeling, this means that the pseudo label is selected from the top 5 candidate valuations. %
For constraint loss, the top 5 candidates are split into valid ($\vval$) and invalid ($\vinval$) valuations and the constraint loss penalizes the invalid valuations.

Results for \textit{pseudo labeling} are presented in Table~\ref{tab:pseudo_labeling_results} and for \textit{constraint loss} in Table~\ref{tab:constraint_loss_results}. Additional plots are provided in Appendix~\ref{app:constraint_satisfaction}. Each experiment was conducted 3 times with different seeds, we show averaged results. 
The results indicate that constraints are beneficial only when there is enough supervised data available. In cases where labeled data is scarce, the addition of unsupervised data can actually harm model performance, suggesting that constraints require a solid supervised foundation to be effective. Furthermore, there is no universally best method across all datasets and supervision ratios. For example, in the medical dataset, constraint loss leads to a 5\% improvement in accuracy when the supervision ratio is 0.5, but decreases by 3\% when the supervision ratio is 0.3. This variability highlights the method's sensitivity to data distribution. 
Additionally, constraint loss is generally less stable, particularly when supervision is minimal. A striking case is observed for constraint loss for the Genbase dataset, where accuracy drops by 70\% with low supervision. However, this approach also provides the largest improvement of 5\% for the Medical dataset. 
In scenarios where a large dataset is unsupervised, pseudo-labeling is often the better choice, as it tends to be more stable and conservative. In contrast, constraint loss is more effective when a significant portion of the dataset is supervised, and it can provide a more substantial performance boost.

\begin{table}[htb]
\centering
\caption{Accuracy difference in percentage between Pseudo Labeling and No Unsupervised Learning. '--' indicates no change.}
\label{tab:pseudo_labeling_results}
\addtolength{\tabcolsep}{-0.1em}
\begin{tabular}{l r r r r r r}
\toprule
Dataset & 0.1 & 0.3 & 0.5 & 0.7 & 0.9 & 0.95 \\
\midrule
Arts & 0.31\% & 0.31\% & -0.33\% & -1.27\% & 0.13\% & -0.33\% \\
Business & -0.16\% & 0.44\% & -0.44\% & 0.64\% & -0.18\% & -0.71\% \\
Cal500 & -- & -- & -- & -- & -- & -- \\
Emotions & 0.17\% & 0.33\% & -0.33\% & -- & 0.66\% & -1.84\% \\
Enron & 0.92\% & 0.17\% & -0.75\% & -0.12\% & -1.24\% & 0.06\% \\
Genbase & 0.50\% & 0.17\% & -0.50\% & 0.50\% & 0.33\% & -4.35\% \\
Image & 0.44\% & -- & 0.56\% & -0.39\% & -0.78\% & -2.28\% \\
Medical & 1.03\% & -1.39\% & -2.58\% & -3.10\% & -0.98\% & -2.53\% \\
Scene & -0.33\%& 0.11\% & -0.47\% & -0.17\% & -1.34\% & -0.17\% \\
Science & 1.82\% & 0.49\% & 0.33\% & -1.09\% & -1.09\% & -0.93\% \\
Yeast & 0.11\% & -0.58\% & 1.09\% & 0.39\% & -0.33\% & -1.32\% \\
\bottomrule
\end{tabular}
\end{table}

\begin{table}[htb]
\centering
\caption{Accuracy difference in percentage between Constraint Loss and No Unsupervised Learning. '--' indicates no change.}
\label{tab:constraint_loss_results}
\addtolength{\tabcolsep}{-0.1em}
\begin{tabular}{l r r r r r r}
\toprule
Dataset & 0.1 & 0.3 & 0.5 & 0.7 & 0.9 & 0.95 \\
\midrule
Arts & 0.91\% & 0.42\% & 0.16\% & -0.62\% & 2.11\% & -0.27\% \\
Business & 0.40\% & 0.02\% & 0.07\% & -1.00\% & 0.09\% & -0.80\% \\
Cal500 & -- & -- & -- & -- & -- & -- \\
Emotions & 0.17\% & -0.83\% & 0.99\% & -0.66\% & 0.33\% & 0.50\% \\
Enron & -0.52\% & 0.81\% & -0.86\% & -0.52\% & -1.27\% & 0.17\% \\
Genbase & 0.50\% & -0.84\% & -- & -1.84\% & -70.52\% & -26.47\% \\
Image & 0.61\% & 0.44\% & -0.17\% & -1.00\% & -0.39\% & -0.44\% \\
Medical & -2.53\% & -3.26\% & 5.43\% & -3.98\% & -5.01\% & -12.40\% \\
Scene & -0.22\% & 0.61\% & -0.28\% & -0.59\% & -0.81\% & -1.31\%\\
Science & 1.58\% & 1.04\% & 1.98\% & 0.09\% & 2.31\% & 1.60\% \\
Yeast & 0.55\% & -1.05\% & 0.69\% & -0.11\% & -0.80\% & 2.25\% \\
\bottomrule
\end{tabular}
\end{table}

\subsection{Impact of the Size of the Base Model}
In this experiment, we evaluate the performance of the $\base$, $\ccn$, and $\baseseq$ models under varying configurations of the $\base$ model. Specifically, we fix the hyperparameters for the sequential model 
and vary the number of layers %
for $\base$.

Table~\ref{tab:different_base} in Appendix~\ref{app:basemodelsize} shows the results.
In general, increasing the number of layers improves the accuracy of both the $\base$ and $\ccn$ models; however, beyond a certain point (5, 6 layers), the performance begins to decrease due to overfitting. In contrast, the $\baseseq$ model consistently outperforms the others and demonstrates greater stability in different configurations. This shows that the improvement brought by the sequential integrator model is not due to its extra parameters, but rather to the better architecture.

\section{Conclusion} \label{sec:conc}

In this work we performed preliminary experiments with a two-stage architecture, where integration is provided by a very expressive model. Since exact inference in the expressive model is intractable, we resorted to heuristic search and sampling techniques.
Our results show that these models are competitive with some prior approaches on some datasets, despite the intractability of evaluation. We also found that the models can learn complex constraints. Perhaps related to this, we do not find a consistent significant advantage for training with constraints. 

One caution is that our experiments are against a limited number of competitors, and in particular we do not compare comprehensively to approaches that assume independence  \cite{semanticloss}, and to restricted expressiveness models \cite{scalingtractable}. In addition they are only run on the tabular datasets of \cite{ccn}, chosen because they are small enough to allow extensive variation without enormous time or computation resources. But these datasets have a number of limitations: the constraints are often very simple, and the amount of data is also limited.
It will be important to look at  datasets with more elaborate constraints and more data, particularly those in the video recognition challenge presented in  \cite{roadr}. While \cite{roadr} naturally lends itself to the two-phase architecture we study here, it  involves labelling of bounding boxes in video frames, where constraints apply universally to bounding boxes. Since base models will disagree on the bounding boxes, comparison metrics are more complex.

Our experiments focus on our implementation of a sequential integrator, and other configurations are also possible. For example, one could consider providing the conditional model only with the current marginal probability, or also with the original input. One could also use stateful conditional models. More work is required to fully understand the implications of our architectural decisions.

\begin{acks}
For the purpose of Open Access, the authors have applied a CC BY public copyright license to any Author Accepted Manuscript (AAM) version arising from this submission. This work has been supported by the Hungarian Artificial Intelligence National Laboratory (RRF-2.3.1-21-2022-00004), the ELTE TKP 2021-NKTA-62 funding scheme. This research was conducted as part of the \href{https://r-ai.co/ukraine}{RAI for Ukraine} program, run by the Center for Responsible AI at New York University in collaboration with Ukrainian Catholic University in Lviv.
\end{acks}

\bibliographystyle{splncs04} 
\bibliography{paper.bib}

\clearpage
\appendix
\onecolumn

\section{The Impact of the Sequential Architecture on a toy example} \label{app:toyexample}

We illustrate the power of the sequential 
architecture with respect to $\ccn$ on a small  problem. In this problem the input consists of two real variables from $[0,1]$ and there are two output variables $O_1$ and $O_2$. There is a single constraint: $O_1 \implies O_2$.
Following \cite{ccn}, we examine three distinct cases based on the relationship between the two output variables: \textit{complete overlap}, \textit{partial overlap}, and \textit{disjoint} targets. We compare extremely small $\ccn$ and Seq-only models, each having 6 neurons and 2 hidden layers. In this experiment, the learning rate is $0.01$, training is performed for $20000$ epochs with a patience of $500$ epochs on  the validation loss. Loss was minimized using
Adam optimizer ($\beta_1$ = 0.9, $\beta_1$ = 0.999). The dataset contains $10{,}000$ samples, uniformly drawn from the input space $[0, 1]^2$, and is split into training, validation and test sets ($35/15/50$).

In this experiment we also include a sequential-only model, a variant mentioned in the body of the paper:

\begin{definition}[Seq-only Model]
\label{def:seqonly}
    A \emph{seq-only} architecture $\seqonly_{\condmodel}$ merges a base and a sequential model into a single neural predictor that relies on the repeated invocation of prefix conditional model $\condmodel$ working directly on input valuations. In this architecture $\condmodel$ has signature $(\val(\vec I), \val(\vec O)) \rightarrow \mathbb{R}$.
\end{definition}

\begin{figure}[htb]
    \centering
    \includegraphics[width=0.46\linewidth]{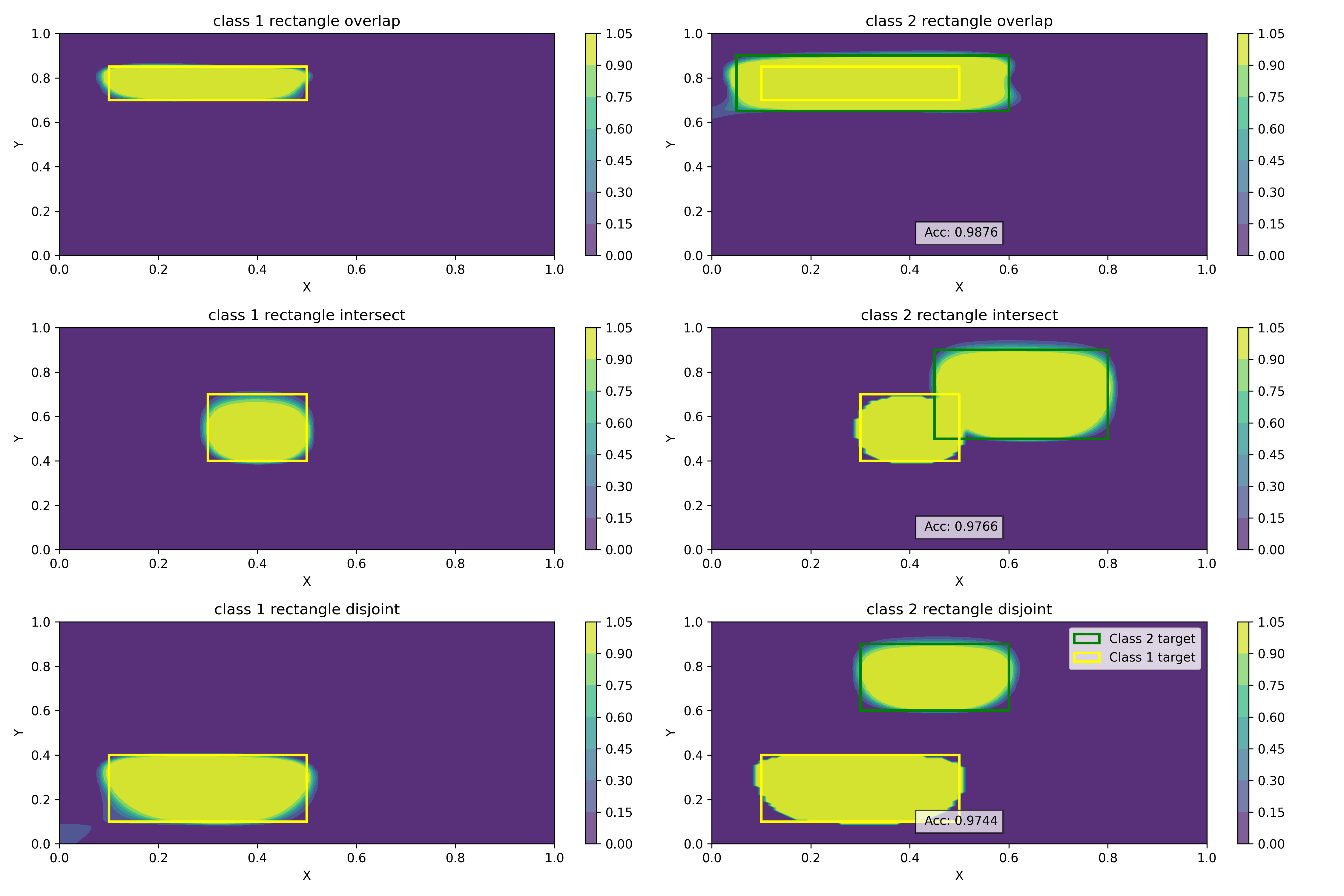}\qquad
    \includegraphics[width=0.46\linewidth]{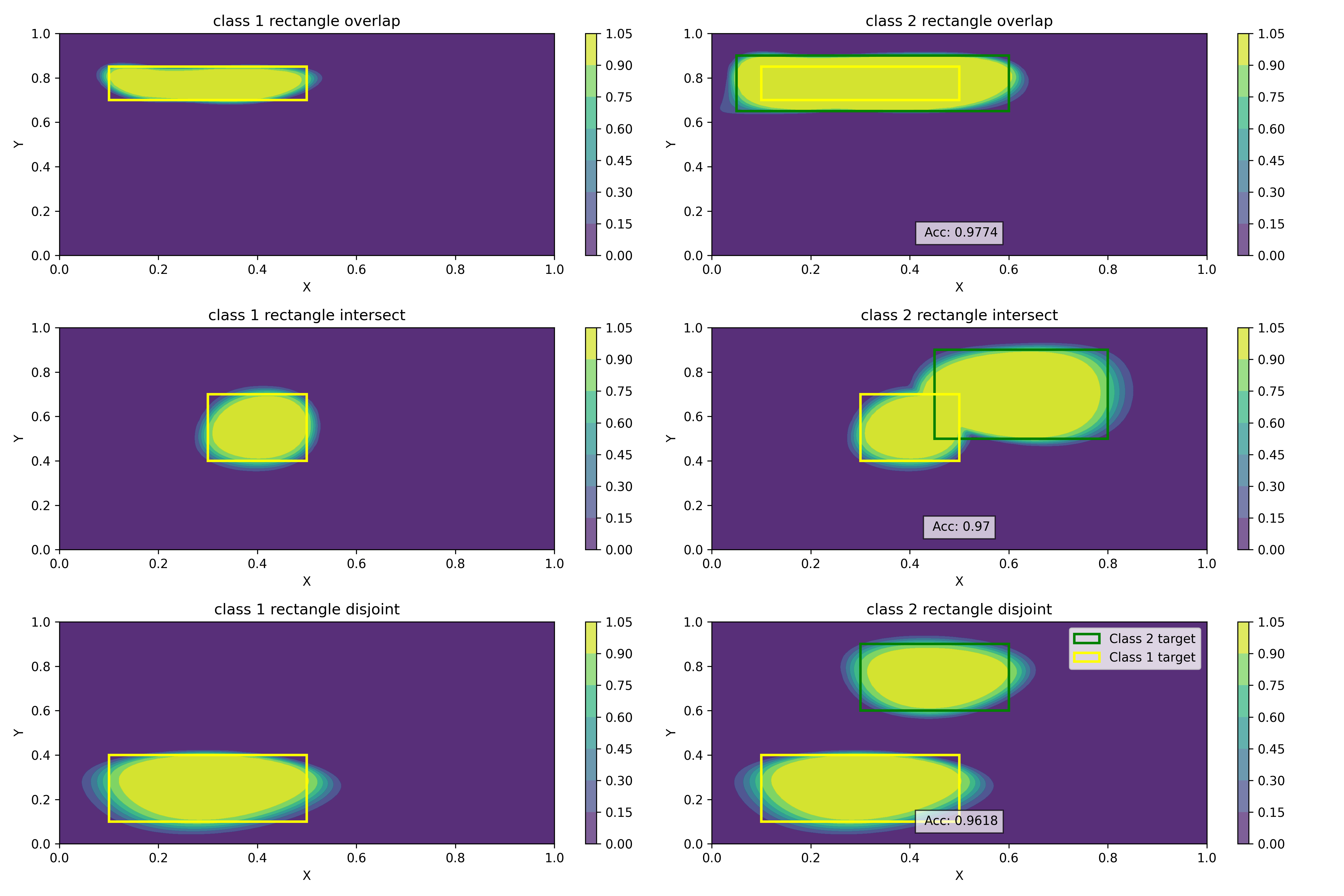}
    \caption{Comparing a Seq-only (left) model with $\ccn$ (right) on the small syntactic problem. In each subplot, the first column corresponds to $O_1$ and the second to $O_2$. Green and yellow rectangles show the grand truth. For each input, color indicates the models probability estimate for the given class. Rows represent different target variable scenarios: \textit{complete overlap}, \textit{partial overlap}, and \textit{disjoint}.}
    \label{fig:combined_figures}
\end{figure}

Figure~\ref{fig:combined_figures} visualizes both the target classes and the learned predictions. With respect to accuracy, the Seq-only model outperforms $\ccn$ in all three scenarios: complete overlap (98,76\% vs. 97,74\%), partial overlap (97,66\% vs. 97\%), disjoint (97.44\% vs. 96.18\%). Note that this is despite the fact that $\ccn$ explicitly ensures constraint satisfaction during inference, while no such guarantee is provided by the $\seqonly$ model.

The $\seqonly$ model performed quite poorly on the more complex datasets, and so we do not include it in tables within the body of the paper.

\section{Constraint Satisfaction with Unsupervised Learning}
\label{app:constraint_satisfaction}

Recall that Tables \ref{tab:pseudo_labeling_results} and \ref{tab:constraint_loss_results} in the body show how training on unsupervised data influences the model's accuracy, with respect to two different constraint-aware training methods. We complement this analysis in Tables \ref{tab:pseudo_labeling_results_violation} and \ref{tab:constraint_loss_results_violation},  showing how constraint violation percentage changes with respect to the mix of supervised and unsupervised data. The changes in constraint violation are small and we see no clear pattern with respect to how it is related to the amount of unsupervised data. 

\begin{table}[htb]
\centering
\caption{Constraint violation percentage difference between Pseudo Labeling and No Unsupervised Learning. Positive numbers indicate weaker performance. '--' indicates no change.}
\label{tab:pseudo_labeling_results_violation}
\addtolength{\tabcolsep}{-0.1em}
\begin{tabular}{l r r r r r r}
\toprule
Dataset & 0.1 & 0.3 & 0.5 & 0.7 & 0.9 & 0.95 \\
\midrule
Arts & -- & -- & -- & -- & -- & -- \\
Business & 0.04\% & -0.09\% & 0.07\% & -- & -- & -- \\
Cal500 & -- & -- & -- & -- & -- & -- \\
Emotions & -- & -- & -- & -- & -- & -- \\
Enron & -- & -- & -0.06 & -- & -- & -0.06\% \\
Genbase & 0.50\% & -- & 0.34\% & 0.50\% & 0.67\% & 0.17\% \\
Image & -- & 0.06\% & 0.06\% & 0.06\% & -- & -- \\
Medical & 0.05\% & -- & -- & -- & -- & -- \\
Scene & -0.06\% & -- & -- & -0.14\% & -0.53\% & -0.81\% \\
Science & -- & -0.04\% & -0.02\% & -- & -- & -- \\
Yeast & -- & -- & -- & -- & -- & -1.38\% \\
\bottomrule
\end{tabular}
\end{table}

\begin{table}[htb]

\centering
\caption{Constraint violation percentage difference between Constraint loss and No Unsupervised Learning. Positive numbers indicate weaker performance. '--' indicates no change.}
\label{tab:constraint_loss_results_violation}
\addtolength{\tabcolsep}{-0.1em}
\begin{tabular}{l r r r r r r}
\toprule
Dataset & 0.1 & 0.3 & 0.5 & 0.7 & 0.9 & 0.95 \\
\midrule
Arts & -- & -- & -- & -- & -- & -- \\
Business & -- & -0.11\% & -- & -- & -- & -- \\
Cal500 & -- & 1.10\% & -- & -- & 4.86\% & -- \\
Emotions & -- & -- & -- & -- & -- & -- \\
Enron & -- & -- & -0.06\% & -- & 0.06\% & -0.06\% \\
Genbase & -- & -- & 0.17\% & 0.34\% & 0.67\% & -- \\
Image & -- & -- & -- & -- & -- & -- \\
Medical & -- & -- & -- & -- & -- & -- \\
Scene & -0.06\% & -- & -- & 0.11\% & -0.56\% & -0.81\% \\
Science & 0.09\% & -0.04\% & -0.02\% & -- & -- & -- \\
Yeast & -- & -- & -- & -- & -- & -1.38\% \\
\bottomrule
\end{tabular}
\end{table}

\newpage

\section{Changing the Size of the Base Model}
\label{app:basemodelsize}

Table~\ref{tab:different_base} summarizes experiments studying how the size of the base model affects model accuracy. We find that adding more layers can help both the base model and $\ccn$, but performance quickly saturates and starts to deteriorate. Hence, the improvement brought by the sequential integrator model in the Base-Seq model is not due to its extra parameters, but rather to the better architecture.

\begin{table}[H]
\centering
\caption{Accuracy comparison of Base, Base-seq (beam), and $\ccn$ models for different number of layers in the Base model on Emotions, Yeast, and Arts datasets. The numbers are averages of 3 runs with different random seeds.}
\label{tab:different_base}
\setlength{\tabcolsep}{3pt} 
\addtolength{\tabcolsep}{-0.1em}
\begin{tabular}{llccccc}
\toprule
\textbf{Dataset} & \textbf{Model} & \textbf{2} & \textbf{3} & \textbf{4} & \textbf{5} & \textbf{6} \\
\midrule
Emotions & Base-seq (beam) & \textbf{0.356} & \textbf{0.356} & \textbf{0.328} & \textbf{0.249} & \textbf{0.271} \\
         & Base            & 0.269 & 0.297 & 0.226 & 0.097 & 0.083 \\
         & CCN             & 0.310 & 0.328 & 0.274 & 0.145 & 0.153 \\
\midrule
Yeast   & Base-seq (beam) & \textbf{0.233} & \textbf{0.236} & \textbf{0.232} & \textbf{0.216} & \textbf{0.201} \\
        & Base            & 0.182 & 0.193 & 0.208 & 0.192 & 0.184 \\
        & CCN             & 0.172 & 0.187 & 0.198 & 0.189 & 0.168 \\
\midrule
Arts    & Base-seq (beam) & \textbf{0.366} & \textbf{0.358} & \textbf{0.341} & \textbf{0.299} & \textbf{0.267} \\
        & Base            & 0.204 & 0.251 & 0.290 & 0.265 & 0.190 \\
        & CCN             & 0.222 & 0.250 & 0.258 & 0.248 & 0.191 \\
\bottomrule
\end{tabular}
\end{table}

\

\end{document}